\documentclass{article}

\usepackage[final]{nips_2017}
\usepackage{natbib}
\usepackage{graphicx}
\usepackage[utf8]{inputenc} 
\usepackage[T1]{fontenc}    
\usepackage{hyperref}       
\usepackage{url}            
\usepackage{booktabs}       
\usepackage{amsfonts}       
\usepackage{nicefrac}       
\usepackage{microtype}      

\usepackage{xcolor}
\usepackage{amsmath}
\usepackage{amssymb}
\usepackage[ruled,vlined]{algorithm2e}
\usepackage{titlesec}
\titlespacing\section{0pt}{8pt plus 2pt minus 2pt}{0pt plus 2pt minus 2pt}
\titlespacing\subsection{0pt}{12pt plus 4pt minus 2pt}{0pt plus 2pt minus 2pt}
\titlespacing\subsubsection{0pt}{12pt plus 4pt minus 2pt}{0pt plus 2pt minus 2pt}

\def\xadv{$\mathbf{x}_{adv}$}

\title{Did you hear that? Adversarial Examples Against Automatic Speech Recognition}
\author{
	Moustafa Alzantot, Bharathan Balaji, Mani Srivastava\\
  Department of Computer Science\\
   University of California, Los Angeles \\
   Los Angeles, CA 90095 \\
  \texttt{\{malzantot,bbalaji,mbs\}@ucla.edu} \\
}

\begin{document}

\maketitle

\vspace{-2em}
\begin{abstract}
\vspace{-1em}
Speech is a common and effective way of communication between humans, and modern consumer devices such as smartphones and home hubs are equipped with deep learning based accurate automatic speech recognition to enable natural interaction between humans and machines. 
Recently, researchers have demonstrated
powerful attacks against machine learning models that can fool them to produce incorrect results. However, nearly all previous research in adversarial attacks has focused on image recognition and object detection models.  In this short paper, we present a first of its kind demonstration of adversarial attacks against speech classification model. Our algorithm performs \textit{targeted} attacks with \textit{87}\% success by adding small background noise without having to know the underlying model parameter and architecture. Our attack only changes the least significant bits of a subset of audio clip samples, and the  noise does not change 89\% the human listener's perception of the audio clip as evaluated in our human study.
\end{abstract}

\section{Introduction}
Recent progress in machine learning and artificial intelligence is shaping the way we interact with
our everyday devices. 
Speech based interaction is one of the most effective means and is widely used in personal assistants of smartphones (e.g. Siri, Google Assistant). These systems rely on running speech classification model to recognize the user’s voice commands.  Although traditional speech recognition models were based on hidden markov models (HMMs), deep learning models are currently the state of art for automatic speech recognition (ASR) ~\cite{graves2013speech},~\cite{amodei2016deep} and speech generation~\cite{oord2016wavenet}.
Despite their outstanding performance accuracies 
in many applications, recent research has shown that neural networks are easily fooled by malicious attackers who can force the model to produce wrong result or to even generate a targeted output value. This kind of attack known as adversarial examples has been demonstrated with high success against image recognition, and object detection models. However, to the best of our knowledge there have been no successful equivalent attacks against automatic speech recognition (ASR) models. 

In this paper, we present an attack approach that fools neural-network-based speech recognition model. Similar to adversarial example generation for images, the attacker will perturb benign (\textit{correctly classified}) audio files by adding a small amount of noise to cause the ASR model to mis-classify or produce a specific target output label. The added noise is small and will be observed by a human listening to the attack audio clip as background noise and will not change how a human recognizes the audio file. However, it will be sufficient to change the model prediction from the true label to another target label chosen by the attacker.

Existing methods for adversarial examples generation such as FGSM~\cite{goodfellow2014explaining}, Jacobian-based Saliency Map Attack~\cite{papernot2016limitations}, DeepFool~\cite{moosavi2016deepfool}, and Carlini~\cite{carlini2017towards} depend on computing the gradient of some output of the network with respect to its input in order to compute the attack noise. For example, in the FGSM~\cite{goodfellow2014explaining} the adversarial noise is computed as:
\[
 \mathbf{x}_{adv} = \mathbf{x} + \epsilon \, sign(\nabla_{\mathbf{x}} J(\theta, \mathbf{x}, y))
\]
The gradient needed to compute adversarial noise can be efficiently computed using backpropagation \textit{assuming attacker knows model architecture and parameters}.  However, backpropagation, \textit{being based on the chain rule}, requires the ability to compute the derivative of each network layer output with respect to the layer inputs. While it is easily done in image recognition models where all layers in the pipeline are differentiable, it becomes problematic to apply same techniques for speech recognition models as they rely on the Mel Frequency Cepstral Coefficients (MFCCs) as features of the input audio data. Therefore, the first layers of an ASR model typically pre-process the raw audio by computing the spectrogram and the MFCC inputs. These two layers are not differentiable and there is no efficient way to compute the gradient through them. While the training process of the neural network does not require backpropagation because MFCCs are considered as model inputs, the generation of adversarial examples would require the gradient. Therefore, gradient-based methods~\cite{goodfellow2014explaining,papernot2016limitations,carlini2017towards,moosavi2016deepfool}) to generate adversarial noise are not directly applicable to speech recognition models based on MFCCs.


Our algorithm generates adversarial noise to perform targeted attacks on ASR. To avoid computing MFCC derivatives, we use a genetic algorithm which is a gradient-free optimization method. Our genetic algorithm based method does not require knowledge of the victim model architecture or parameters and can be used for black box attacks without training substitutive models. We evaluate our attack using the speech commands recognition model~\cite{sainath2015convolutional} and the speech commands dataset~\cite{speechcommands}. Our results show that targeted attacks succeed 87\% of the time while adding noise to only the 8 least-significant-bits of a subset of samples in a 16 bits-per-sample audio file.  We evaluate the effect of noise on human perception of the audio clip with a user study. Results show that the noise did not change the human decision in 89\% of our samples and listeners still recognize the audio as its original label.

\textbf{Adversarial Attacks on Audio: }
\begin{figure*}[!t]
\centering
\includegraphics[scale=0.30]{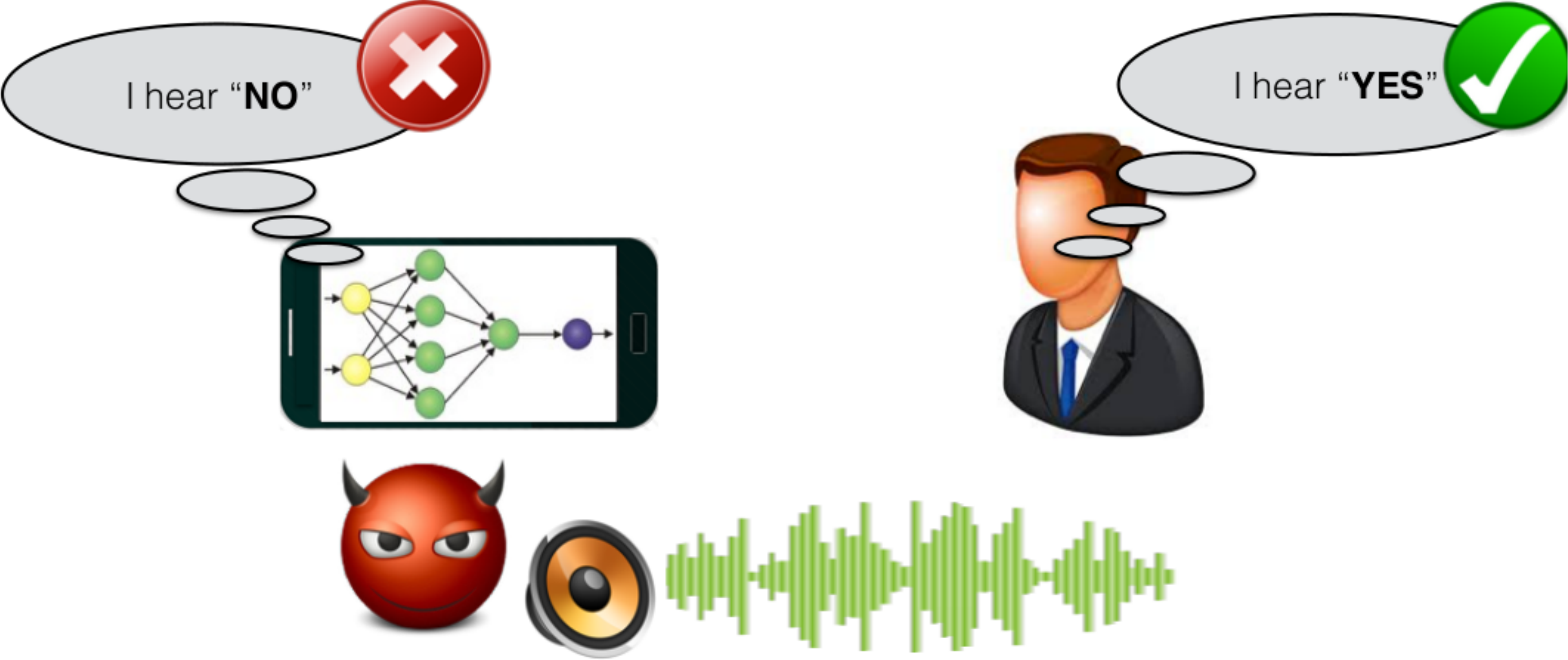}
\caption{Adversarial attacks on speech commands: a malicious attacker adds small noise to the audio such that it is misclassified by the speech recognition model but does not change human perception.}
\vspace{-2mm}
\label{fig:adv_speech}
\end{figure*}
Adversarial examples refer to inputs that are maliciously crafted by an attacker to fool machine learning models. Adversarial examples are typically generated by adding noise to the inputs that are correctly classified by the model, and the added noise should be imperceptible for humans. 
To create adversarial examples for speech recognition models an attacker takes a \textit{legitimate} audio file perturbs it by adding an \textit{imperceptible} noise that causes the machine learning speech recognition model to mis-classify the input \textit{and possibly produce a desired target label}. We demonstrate this in Figure ~\ref{fig:adv_speech}, where the attacker adds noise to an audio clip of the word "YES" that the machine learning model classifies as "NO" while the human still recognizes as "YES".

\textbf{Prior Audio Attacks: } While recent research uncovered potential attacks against speech recognition models, 
the demonstrated attacks do not represent an instantiation of adversarial examples~\cite{goodfellow2014explaining} as witnessed with image recognition models. Backdoor~\cite{roy2017backdoor} exploits the non-linearities of microphones in smart devices to play audio at a frequency that is inaudible to humans (40 kHz), 
but creates a shadow in the audible range of the microphone. Backdoor harnesses this phenomenon to 
block microphone in places such as movie theaters. However, the attack requires an array of specialized high frequency speakers. DolphinAttack~\cite{song2017inaudible} exploits the same non-linearities in microphones to create commands audible to speech assistants but inaudible to humans. Notably, in both methods~\cite{roy2017backdoor,song2017inaudible} the attack sound is not heard by the human at all, while an adversarial example should be recognized by a human as benign while misclassified by the speech recognition model.
The attack to closest adversarial examples is the ``Hidden Voice Commands'' by Carlini et al.~\cite{carlini2016hidden} that generates sounds that are unintelligible to human listeners but interpreted as commands by devices. Nevertheless, it does not represent an adversarial attack because the samples they generate are aimed to be `unrecognizable' by humans, but it can still lead to suspicion.  A more stealthy and powerful attack will maintain the listener interpretation of the attack samples as something benign. 

\textbf{Threat Model: }
Our attack assumes a black-box threat model where the attacker knows nothing about the model architecture and parameter values, but is capable of querying the model results. Precisely, the victim model is used by the attacker as a black box function $f(\mathbf{x})$ while mounting his attacks. Such that: $
 f: \mathbb{X} \longrightarrow [0,1]^K$
where $X$ is the space of all possible input audio files, and the output $[0, 1]^K$ represent the prediction probability scores to each one of the possible $K$ output labels. The output values are obtained from the final \texttt{Softmax} layer commonly used in classification models.

\section{Generating Adversarial Speech Commands}
\label{gen_inst}
\begin{algorithm}[!h]
\IncMargin{1.5em}
\DontPrintSemicolon
\SetAlgoLined
\SetKwInOut{Input}{Inputs}\SetKwInOut{Output}{Output}
\SetKwFunction{InitializePopulation}{InitializePopulation}
\SetKwFunction{Crossover}{Crossover}
\SetKwFunction{ComputeFitness}{ComputeFitness}
\SetKwFunction{Mutate}{Mutate}
\SetKwData{Pop}{population}
\SetKwData{NextPop}{Next population}
\Input{Original benign example $\mathbf{x}$\\target classification label $t$}
\Output{Targeted attack example $\mathbf{x}_{adv}$}
\BlankLine
\tcc{Initialize the population of candidate solutions}
\Pop $\longleftarrow$ \InitializePopulation{$\mathbf{x}$}\;
iter\_num = 0\;
\While{$iter\_num < max\_iter$} {
scores $\longleftarrow$ \ComputeFitness{\Pop}\;
$\mathbf{x}_{adv} \longleftarrow $ \Pop$[ argmax(scores)]$\;

\If{argmax $f(\mathbf{x}_{adv}) = t$} {
	break \tcp{Attack succeeded, Stop early. }
}
\tcc{Compute selection probabilities.}
$select\_probs \longleftarrow Softmax(\frac{scores}{Temp})$\\
\NextPop $\longleftarrow \{\, \}$ \;
\For{$i \leftarrow 1 $ \KwTo $size$}{
	Select $parent_1$ from \Pop according to probabilities $select\_probs$\;
	Select $parent_2$ from \Pop according to probabilities $select\_probs$\;
	child = \Crossover{$parent_1$, $parent_2$}\;
	\NextPop = \NextPop $\bigcup \{ child \}$\;
}
\lForEach{child of \NextPop}{\Mutate{$child$}}
\Pop $\longleftarrow$ \NextPop\;
$iter\_num = iter\_num + 1$\;
}
\KwRet{\xadv{}}\;
\caption{Generation of Targeted Adversarial Audio Files using Genetic Algorithm}
\label{alg:main}
\end{algorithm}

We use gradient free genetic algorithm based approach to generate our adversarial examples as shown in \ref{alg:main}. The algorithm accepts an original benign audio clip $\mathbf{x}$ and a target label $t$ as its inputs. It creates a population of candidate adversarial examples by adding random noise to a subset of the samples within the given audio clip. To minimize the noise effect on human perception, we add noise to only least-significant bits of a random subset of audio samples. We compute fitness score to each population member based on the prediction score of the target label and produce the next generation of adversarial examples from the current generation by applying selection, crossover and mutation. Selection means that population members with higher fitness value are more likely to become part of the next generation. Crossover takes pairs of population members and mixes them to generate a new `child' that will be added to the new population. Finally, mutation adds random noise with very small probability to the child before passing it to the future generation. The algorithm iterates on this process for preset number of epochs or until the attack is found successful.

Due to space constraints, we omit the detailed description of some subroutines and hyper-parameters used in our algorithm. To assist other researchers to reproduce our results, we have made our implementation (with the same hyper-parameter values used for evaluation results reported in this paper) available at \url{https://git.io/vFs8X}.

\section{Evaluation}
\textbf{Speech Recognition Model: }
We evaluate our attack against the Speech Commands classification model~\cite{sainath2015convolutional} implemented in the TensorFlow~\cite{abadi2016tensorflow} software framework. This model is an efficient and light-weight keyword spotting model based on convolutional neural network and achieves 90\% classification accuracy on the speech commands~\cite{speechcommands} dataset. 
The speech commands dataset~\cite{speechcommands} is a crowd-sourced dataset consisting of 65,000 audio files. Each file is a one second audio clip of single words like: "yes", "no", "up", "down", "left", "right", "on", "off", "stop", or "go".

\textbf{Targeted Attack Results: }
For the targeted attack experiment, we randomly select 500 audio clips from the dataset at 50 clips per labels (after we exclude the "silence" and "unknown" labels). We produce adversarial examples from each file such that it will be classified as a different target label. For example, for an audio clip of "yes", we produce adversarial examples that are targeted to be classified as "no", "up", "down", "left", etc. This means for input audio clip we produce 9 adversarial examples leading to a total count of 4500 output files. Samples of our targeted attack output can be listened to at \url{https://git.io/vFs42}.
\begin{figure}[!h]
\centering
\includegraphics[scale=0.45]{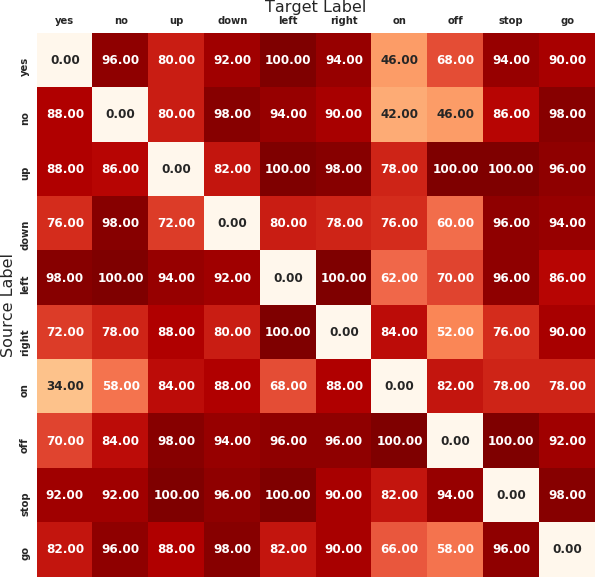}
\caption{Percentage of success for every (source, target) targeted adversarial attack.}
\label{fig:targeted_results}
\end{figure}
Figure~\ref{fig:targeted_results} shows the result of our targeted attack. Our algorithm was successful 87\% in performing targeted adversarial attack between any source-target pair.  
We limit the number of iterations in our algorithm to 500. If the algorithm fails to find a successful targeted attack within 500 iterations, we declare this as failure. The median time to generate an adversarial audio file is 37 seconds on a desktop machine with Nvidia Titan X GPU. A more successful attack can be possible if we increase the limit of noise or number of iterations.

\begin{table}[!b]
\centering
\begin{tabular}{|c|c|c|}
\hline
Attack labeled as source & Attack labeled as target & Attack labeled as other\\
\hline
89\% & 0.6\% & 9.4\% \\
\hline 
\end{tabular}
\caption{Human perception of adversarial examples generated by our attack}
\label{tab:human_exp}
\vspace{-1em}
\end{table}

\textbf{Human Perception Results: }

In order to assess the effect of added adversarial noise on human listeners, we conducted a human study where we recruited 23 participants, and we asked them to listen to and label successful adversarial audio clips we generated. In total, the study participants labeled 1500 audio clips. The participants were not told what is the source or target labels of the audio clips they were provided.

Results from our human experiment shown in Table~\ref{tab:human_exp} show that 89\% of participants were not affected by the added noise and they still label the heard audio at the source label while the machine learning model labels all of them as the target label.

\section{Discussion}
In this section, we discuss the limitations and possible future directions for our study.

\textbf{Using MFCC inversion for white box attack:} Our attack algorithm does not require knowing the model architecture or its parameters and it only uses the victim model as a black box. In a white-box scenario where an attacker can utilize his knowledge about victim model, a stronger attack may be possible. However, this approach will face the hurdle of how to do back-propagation through the MFCC and spectrogram layer. One idea is to compute the adversarial noise with respect to the MFCC layer outputs as the classification model inputs, then use MFCC inversion~\cite{boucheron2008inversion} to reconstruct the adversarial audio. Further experiments should be done to evaluate the quality of this approach.

\textbf{Evaluation against a larger ASR model and complete sentence  generation:} An interesting question is if the more powerful state-of-art ASR models are also affected by adversarial examples, and whether we can generate adversarial sentences instead of just adversarial audio clips of single words.

\textbf{Untargeted attacks: } We reported the results of our targeted attacks where the attacker specifies the desired output label. In addition, we achieved 100\% success rate with our untargeted attacks. Although the untargeted attack is considered a weaker type of attack, further study of the untargeted attacks can be useful to study model robustness against adversarial noise.

\textbf{Over the air attack: }
In our evaluation, we assume that the attacker feeds the audio clip directly to the classification model. However, a more realistic and powerful attack will succeed even when we play the adversarial audio clip from the speaker while the victim model picks the audio from the microphone. This is harder to achieve, and we plan to study it in follow-up research.

\subsubsection*{Acknowledgments}
This research was supported in part by the NIH Center of Excellence for Mobile Sensor Data-to-Knowledge (MD2K) under award 1-U54EB020404-01,  the U.S. Army Research Laboratory and the UK Ministry of Defence under Agreement Number W911NF-16-3-0001, and the National Science Foundation under award \# CNS-1705135. Any findings in this material are those of the author(s) and do not reflect the views of any of the above funding agencies. The U.S. and U.K. Governments are authorized to reproduce and distribute reprints for Government purposes notwithstanding any copyright notation hereon. 


\medskip
\small
\bibliographystyle{abbrv}
\bibliography{nips17_deception.bib}

\end{document}